\documentclass{IOS-Book-Article}

\usepackage{mathptmx}
\usepackage{cite}  
\usepackage[pdftex]{graphicx} 
\usepackage{url}

\def\hb{\hbox to 10.7 cm{}}

\begin{document}

\pagestyle{headings}
\def\thepage{}

\begin{frontmatter} 

\title{
A correlational analysis of multiagent sensorimotor interactions:
clustering autonomous and controllable entities
}


\author[A]{\fnms{Mart\'i} \snm{S\'anchez-Fibla}%
\thanks{Corresponding Author: SPECS, Technology Department, Universitat Pompeu Fabra, Carrer de Roc Boronat 138, 08018 Barcelona, Spain. E-mail:
marti.sanchez@upf.edu.}},
\author[A]{\fnms{Cl\'ement} \snm{Moulin-Frier}},
\author[A]{\fnms{Xerxes} \snm{Arsiwalla}}
and
\author[A,B]{\fnms{Paul} \snm{Verschure}}

\address[A]{SPECS, Universitat Pompeu Fabra. Barcelona, Spain.}
\address[B]{ICREA, Instituci{\'o} Catalana de Recerca i Estudis Avan\c{c}ats, Barcelona}

\begin{abstract}
A first step to reach Theory of Mind (ToM) abilities (attribution of beliefs to others) in synthetic agents through sensorimotor interactions, would be to tag sensory data with agent typology and action intentions: autonomous agent X moved an object under the box. We propose a dual arm robotic setup in which ToM could be probed.
We then discuss what measures can be extracted from sensorimotor interaction data (based on a correlation analysis) in the proposed setup that allow to distinguish self than other and other/inanimate from other/active with intentions. 

We finally discuss what elements are missing in current cognitive architectures to be able to acquire ToM abilities in synthetic agents from sensorimotor interactions, bottom-up from reactive agent interaction behaviors and top-down from the optimization of social behaviour and cooperation.
\end{abstract}

\begin{keyword}
Theory of Mind, Sensorimotor Learning, Correlation Matrix
\end{keyword}
\end{frontmatter}
\markboth{May 2017\hb}{May 2017\hb}

\section{Introduction}

Theory of Mind (ToM) is commonly referred as the ability to attribute beliefs and intentions, to others, so being able to understand other people's minds and infer other's mind states. The so-called "Sally-Anne test" is a psychological experiment designed to probe attribution of beliefs, probing ToM abilities, and was first studied in relation to autism \cite{baron1985does}. In the test, the participant is presented with fictional characters Sally with her basket, and Anne with her box. Sally places an object in her basket and leaves the room. Then Anne opens the basket, takes the object and puts it in the box. Sally comes back and the participant is asked, where will Sally look for the object? The test is passed if the basket is selected. Although we know that the object is in the box, Sally cannot know it because she could not see Anne placing it in her box.
Related to ToM is the ability to attribute false beliefs: assuming that people can have beliefs (Sally believes the object is in the basket) and that these can be in non-accordance of one's own beliefs (we know that the object is in the box). Attribution of false beliefs can appear as early as 13 months \cite{surian2007attribution}. 

The question now is: can a synthetic agent running current state of the art artificial controllers or cognitive architectures acquire ToM abilities from sensorimotor interactions with the world (perception and actuation)? 

\begin{figure}[!t]
\centering
\includegraphics[width=4.2cm]{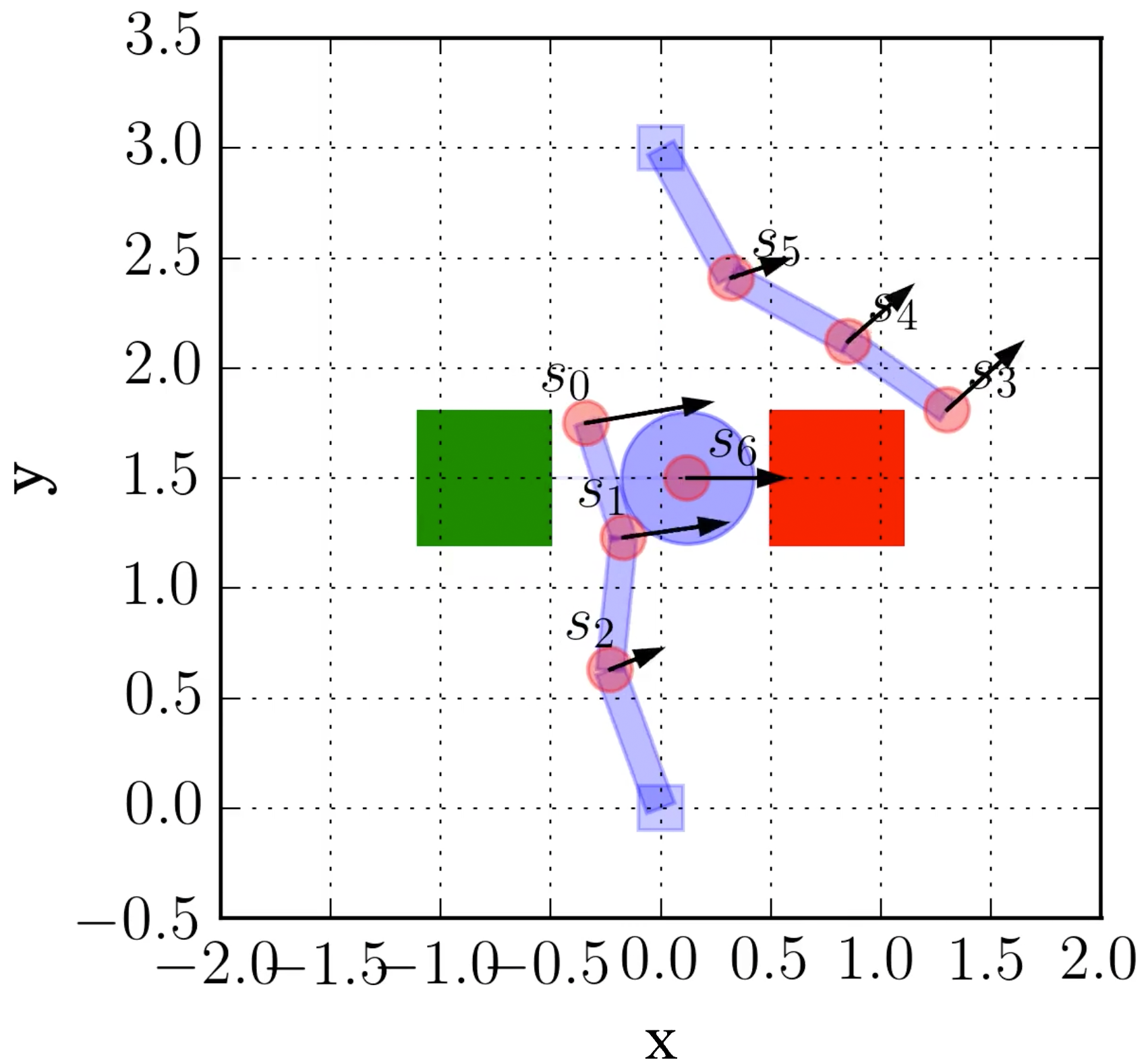} 
\hspace{0.1cm}
\includegraphics[width=7.5cm]{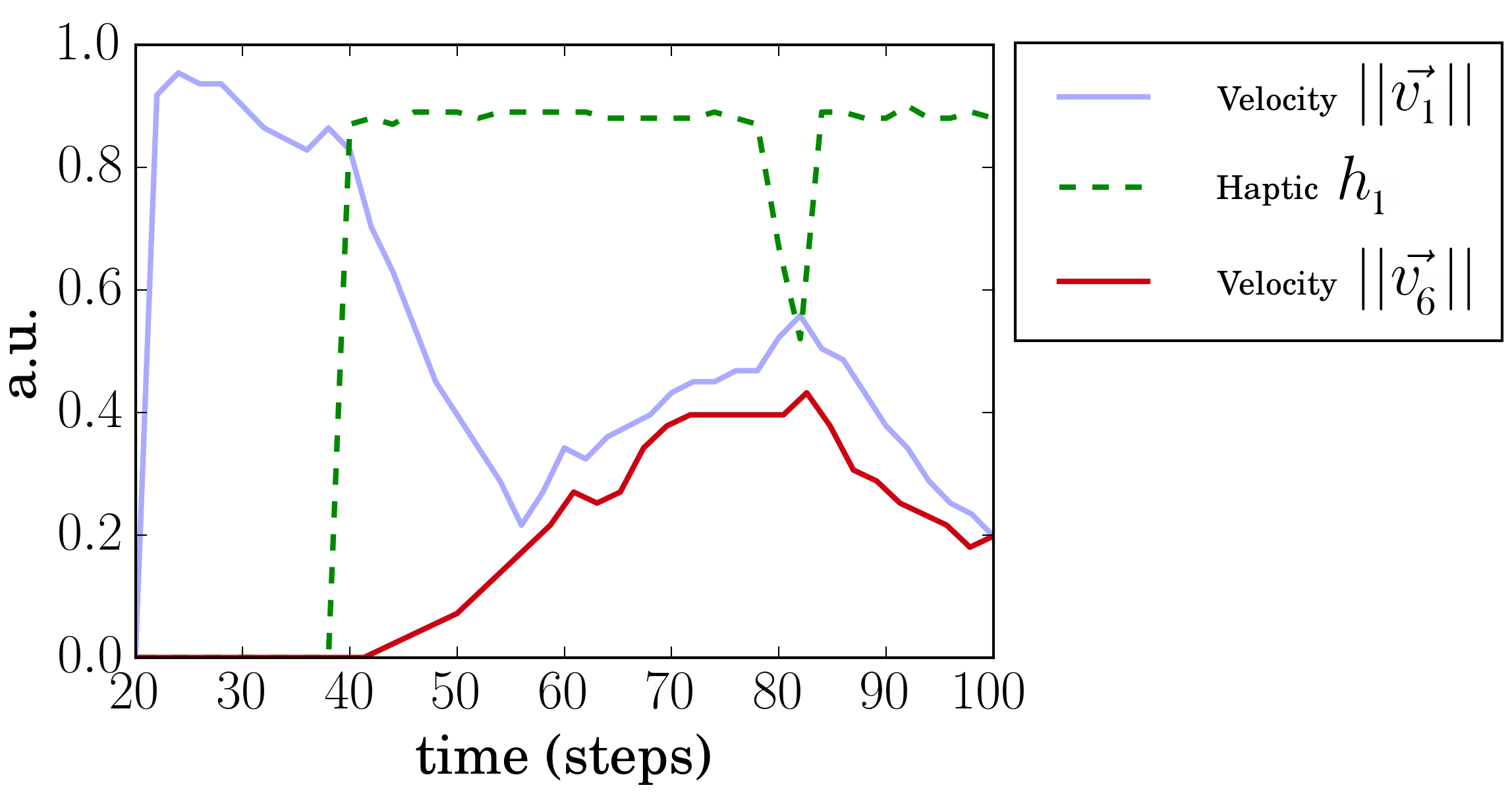} 

\caption{Dual Robotic Simulation Arm Setup. \textbf{Left:} Two 3DOF (arms can be of any DOFs)  robotic arms face each other and, in the middle, an object is constrained to move in a horizontal axis and cannot be seen when under the boxes (indicated by the green and red squares). The vision of agents is driven by the interest points $s_0,...,s_5$ in its joints and in the object $s_6$. Agents also receive tactile feedback when contacts occur. \textbf{Right:} Time series data in a moment of contact of sensory point $s_1$ and $s_6$. The haptic signal (green dashed line) increases suddenly in the moment of contact (at 40 time steps). At this point the velocity of the object increases as well.}  
\label{fig:setup}
\end{figure}

Our positioning statement is that acquiring ToM abilities requires to label and cluster a SensoriMotor Memory (SMM from now on) of the sensory experience according to what can I control (self/other distinction), how are the other entities controlled (passive/active agents distinction, see Section  \ref{structures} for a further explanation on SMMs). We propose in Section \ref{correlation} a correlation analysis that could solve part of the labeling  problem.
It is in this sense that we refer to social sensorimotor contingencies (socSMC) which are the extraction of low-level regularities (contingencies) in sensorimotor data (as defined in \cite{o2001sensorimotor}) that are particular of social interaction.  We evaluate the correlation analysis in a dual robotic arm setup (freely available\footnote{The simulation can be download from \url{https://github.com/santmarti/PythonRobot2DSim}. A video TwoArmSetup.mp4 is also available under the folder videos in the same github repository.}, see Figure \ref{fig:setup} and Section \ref{setup} for details) in which two robot arms facing each other can reach an object that can slide under two boxes, one green box in the left (analogous to Sally's basket) and one red box in the right (analogous to Anne's box). Each robot has access to the whole visual scene through perceived sensory points (see Figure \ref{fig:setup}).   The socSMCs extracted from interacting with the object would be very different than the ones extracted from the same interaction if the other agent is also acting on it.

The final long-term aim of this research is to settle the computational basis of how ToM abilities could arise from low-level sensorimotor interactions, bottom-up from agent interaction behaviors. In this sense we differ from multiagent approaches like \cite{marsella2004psychsim} in which beliefs are symbols added as logical facts and inference is performed through rule systems.  
In a later phase, and building from the bottom up interacitons, ToM abilities could be refined through top-down optimization of social behaviour and cooperation. 

\section{Dual robotic arm setup}
\label{setup}

The so-called ToM setup is composed of two arm robots, an object and two colored boxes that can have no physical interaction with the other elements but occlude them when they move underneath (see Figure \ref{fig:setup}). The two arm robots face each other so that there is an area where both can interact, affecting its own action outcomes. Depending on the current state, each agent can move freely, interact with the object alone, interact with the other agent, interact with the other agent through the object. 

Each entity in the setup has associated sensory points (attached to joints in the robots). All the set of sensory points constitutes the so-called sensory space $S$ and it considers two modalities: visual motion information and haptic inputs. $S_t$ consists of a set of sensory interest points (sensory points from now on): $S=\lbrace s_0, \cdots, s_i, \cdots \rbrace$. We define a sensory point as being a tuple: 
$s_i = \langle p_i = \langle x_i,y_i\rangle, h_i\rangle$ where $p_i$ corresponds to the position in the 2D cartesian space of the sensory point and $h_i$ is the haptic signal sensed at $s_i$. Predefined ranges are $h_i \in [0..1] $ (being $0$ for no contact and $1$ for maximal contact), $||\vec{v_i}|| \in [0..1], x_i \in [-4,4], y_i \in [-2,4]$.

Robots have three joints that can be actuated independently. The motor space of the setup is defined by $M$. At any moment, the motor space consists of the state of each motor joint (the angle $a_i$) and its speed or torque being applied at each of its motors (that we will denote $m_i$): $M = \langle a_0, ..., a_n \rangle$ is the state of all motors, their angles, and  $\Delta M = \langle m_0,...,m_n \rangle$ their torques or velocities, being $n$ the total number of degrees of freedom (DOFs).

\section{Sensorimotor Structures}
\label{structures}

We start discussing that one has to consider three elements of an SMM:  \textit{1)} its structure, components and relations between components \textit{2)} Exploration of SMM spaces: how it is filled: how components are added and how they are linked together \textit{3)} Exploitation of SMM structure: What are the procedures used to query the SMM structure, the queries that it accepts; how action is selected; thus what kind of planning can be done with it.     

\textit{Distributed Adaptive Control (DAC)}:
It is common to store a history of observed sensory elements together with the executed motor commands to generate sequences of sensory-motor couplets. This is the approach used in the Contextual Layer of DAC. A directed-graph network version of this Contextual Layer is presented in \cite{duff2011biologically} in which nodes represent possible occurrences of sensory-motor couplets (also including obtained reward) and these are linked by lateral connectivity through Hebbian learning if two nodes happen to be observed one after the other. 
\textit{1)} The SMM is a directed graph of nodes composed by: sensory, motor, reward states. Initially Nodes are connected randomly.
\textit{2)} Nodes are randomly created. Edges between nodes are reinforced if their nodes subsequently co-occur in time with Hebbian learning.
\textit{3)} The network is driven by the instantaneous sensory input. Thus a number of nodes are activated and activity is propagated through the graph taking into account reward values and learned lateral connectivity. A winner take all procedure selects a node and its corresponding action is executed.

\textit{Graph Sensorimotor Structures}:
Toussaint \cite{toussaint2006sensorimotor} presents a similar SMM directed-graph network approaches. \textit{1)} Nodes are associated to observed sensory states. Edges are experienced sensory state transitions and are labeled by the motor command that produced that sensory change. Nodes can also be added depending on novelty factors via a Hierarchical Self-Organizing Map \cite{butz2010self}. \textit{2)} Edges are created/reinforced when a transition from one node to another that was made executing a particular motor action is observed. \textit{3)} Graphs can be queried with graph algorithms from current sensory nodes to target sensory nodes executing edges in the path.


\vspace{0.3cm}

We claim that state of the art sensorimotor memory (SMM) structures used in cognitive architectures like SOAR \cite{laird2008extending} and DAC \cite{duff2011biologically}\footnote{Subsumption architecture \cite{brooks1986robust} does not define an SMM, but a stack of behaviors that are sensory driven.} or sensorimotor structures defined in purpose like \cite{butz2010self} and \cite{toussaint2006sensorimotor} record and relate sensorimotor states usually in a graph like structure but don't allow to relate those sensory states to belonging to other agents or objects. The sensory information is not processed up to a level where moving agents can be distinguished from passive objects and thus SMM structures cannot record information relating to who did what, or who knows what, a prerequisite for passing the "Sally and Anne" used for probing ToM \cite{baron1985does,surian2007attribution}.



\section{Correlation analysis of sensorimotor data}
\label{correlation}

\begin{figure}[!t]
\centering
\includegraphics[width=12.5cm]{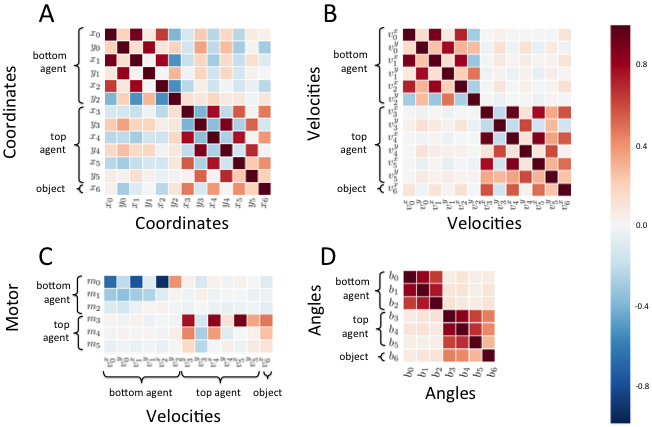}
\caption{Correlation matrices extracted from 1 hour of random sensorimotor interactions. A. Correlation between time series of the coordinates of the sensory points. B. Matrix correlations of the time series of $x$ and $y$ velocity components of the sensory points. C. Correlation of the motor signals of the 3 joins of each agent with the velocities of the sensory points. D. Correlation matrix of the angle of movement of the sensory points. } 
\label{fig:corr}
\end{figure}

Given the low-level strategies adopted by the SMMs described previously, we present a correlation analysis in the direction of identifying the sensory moving points that are controllable by the agent, cluster them in different entities, some being passive and some actively autonomous. 
We record one hour of sensorimotor interaction by random motor movements of both agents.
We show in Figure \ref{fig:corr} different correlation matrices of SMM time series. Each cell in a matrix is the correlation of two time series. In the following list, the components that are considered from the SMM recorded interactions are explained:

\begin{scriptsize}
\begin{itemize}
\item $m_{0..2}$, $m_{3..5}$: motor delta actions of the bottom and top agent.
\item $x_{0..2},y_{0..2}$, $x_{3..5},y_{3..5}$  : $x$ and $y$ coordinates of the sensory points of the bottom and top agents.
\item $x_6,y_6$ : $x$ and $y$ coordinates of the sensory point of the object.
\item $v^x_{0..2},v^y_{0..2}$,$v^x_{3..5},v^y_{3..5}$ : $x,y$ velocity components of the sensory points of the bottom and top agents.
\item $b_{0..2}$,$b_{3..5}$ : angles of movement of the sensory points of the bottom and top agents.
\end{itemize} 
\end{scriptsize}

We observe that patterns of correlations arise between sensory points positions (Figure \ref{fig:corr}A) and a signature of the proximo-distal organization of the arm of each agent is present. The closer two joints are, the larger their correlation is: $corr(x_0,x_1)$ is less than $corr(x_1,x_2)$ for the bottom agent and $corr(x_3,x_4)$ is less than $corr(x_4,x_5)$.
This intra-agent correlation pattern is very similar among the two agents. By matching their own joint position correlation pattern with the one observed from another agent, this could provide a first level of mirroring between agents, where each one is able to match its own kinematic structure with the one of the other. We observe, as well, a correlation of the object coordinate $x_6$ with the second agent: $x_3$, $x_4$ and $x_5$, meaning that it interacted much more with the object than the bottom agent and also revealing that the object can be pushed (reminiscent of the notion of affordance \cite{sanchez2011acquisition}). 

Second, we observe a similar correlation pattern between an agent joint velocities (Figure \ref{fig:corr}B). Interestingly, this pattern is also similar between agents, providing a second level of mirroring, where agents can discover that similar movements of the other provide similar effects on the objects, paving the way to a notion of shared affordances useful for joint action planning. 
The angles of movement of the sensory points are also correlated in between agents  (Figure \ref{fig:corr}D).

Motor signals are correlated with sensory points velocities (Figure \ref{fig:corr}C). The bottom of joint of the bottom agent ($m_0$) is strongly negatively correlated with the $x$ velocity components of sensory points $s_0$, $s_1$ and $s_2$, because this joint moves the whole arm. The same happens with the top agent. This matrix can assess controllability characteristics and could restrict the forward model to be learned to the relevant signals. Again we can observe that the object $x$ velocity component is more correlated with the top agent as it interacted more with it.

An agent could find structure in an apparently random set of sensory points movements by observing the above-mentioned correlations. Extracted from the motor velocities correlations it could assess controllability. From the clustered coordinates and velocities correlation matrices it could extract its own sensory points and assess the similarity of the ones of the other agent. From assessing autonomy of the sensory points it could deduce that the object never moves by itself and can only be moved through movement of another agent. 

\section{Conclusion}
 
We have addressed the fundamental question of the minimum and necessary requirements for an agent to acquire ToM in particular agency segmentation. We have introduced a multi-agent setup in which two arm robots can interact together with an object. From random movements we extract sequences of data based on the dynamics of sensory points attached to the agent body parts and to the object.
We make a preliminary analysis of those interactions in terms of the time series correlations. We find that agents can be distinguished by observing the correlations between their own joints movement data (coordinates and velocities of their own sensory points). Correlation of motor signals and velocites of sensory points reveals controllability of moving entities and restricts the forward model to be learned.

We also discuss how state of the art sensorimotor structures are unable to label data according to controllability and agent identity, and thus are unable to answer questions like: who did what, or what's the state of the world form another agent point of view. An observer must be able to distinguish multiple agents on morphological grounds and segregate them in its maintenance of multiple ToMs. 

Recently, model-free and model-based deep reinforcement learning have been very successful at learning how to control an agent to maximize long-term reward based only on screen pixel values \cite{lake2016building}. Identifying controllability and agent identity and type (passive/active) can facilitate the transfer of knowledge from one domain to another, a feature that is lacking in both model-free and model-based approaches \cite{moulin2017embodied}.

In the future ToM abilities could be quantified in terms of complexity measures that have been used to describe a morphospace of consciousness in \cite{arsiwalla2017morphospace}.

\vspace{.35cm}

\noindent \textbf{Acknowledgments}
\begin{small}
Research supported by INSOCO-DPI2016-80116-P, socSMC-641321—H2020-FETPROACT-2014 and ERC-2013-ADG-341196. We thank anonymous reviewers.
\end{small}  

\vspace{-.3cm}

\bibliographystyle{plain}
\bibliography{socsmc}

\end{document}